\documentclass[journal, twoside]{IEEEtran}
\IEEEoverridecommandlockouts
\usepackage{lipsum}
\usepackage{hyperref}
\usepackage{placeins}
\usepackage{mathtools}
\usepackage{graphicx}
\usepackage{float}
\usepackage{setspace}
\usepackage{color}
\usepackage[dvipsnames]{xcolor}
\usepackage{cite}
\usepackage{indentfirst}
\usepackage{amssymb}
\usepackage{amsmath}
\usepackage{amsbsy}
\usepackage{subfigure}
\usepackage{graphicx}
\usepackage{setspace}
\usepackage{soul}
\usepackage{algorithm}
\usepackage{algorithmic}
\usepackage{threeparttable}
\usepackage{booktabs}
\usepackage{babel}

\begin{document}

\title{An Adaptive Control Algorithm for Quadruped Locomotion with Proprioceptive Linear Legs}

\author{Bingchen Jin\textsuperscript{$\dagger$}, Yueheng Zhou\textsuperscript{$\dagger$}, Ye Zhao, Ming Liu, Chaoyang Song, Jianwen Luo\textsuperscript{$*$}
\vspace{-5mm}

\thanks{Manuscript received: Month date, 20xx; This work was supported in part by National Natural Science Foundation of China under Grant 51905251. \textit{(Corresponding author: J. Luo. Email: luojianwen1123@gmail.com.)}}


\thanks{B. Jin and Y. Zhou are with Shenzhen Institute of Artificial Intelligence and Robotics for Society (AIRS), Shenzhen 518172, China.} 
\thanks{Y. Zhao is with the George W. Woodruff School of Mechanical Engineering, Georgia Institute of Technology, USA.}
\thanks{M. Liu is with Active Vision Group (AVG), Robotics and Multi-perception Lab (RAM-LAB), Department of Electronic and Computer Engineering, Hong Kong University of Science and Technology, Hong Kong, China.}
\thanks{Y. Zhou and C. Song are with the Southern University of Science and Technology (SUSTech), Shenzhen 518055, China.}
\thanks{J. Luo is with the Hong Kong University of Science and Technology (Guangzhou), Guangzhou, 510000, China.}
\thanks{$\dagger$ Bingchen Jin and Yueheng Zhou are co-first authors.}
}

\markboth{SUBMITTED MANUSCRIPT FOR PEER REVIEW}%
{JIN  \MakeLowercase{\textit{et al.}}: High-Payload Online Identification and Adaptive Control for an Electrically-actuated Quadruped Robot}

\maketitle

\begin{abstract}
Quadruped robots manifest great potential to traverse rough terrains with payload. Numerous traditional control methods for legged dynamic locomotion are model-based and exhibit high sensitivity to model uncertainties and payload variations. Therefore, high-performance model parameter estimation becomes indispensable. However, the inertia parameters of payload are usually unknown and dynamically changing when the quadruped robot is deployed in versatile tasks. To address this problem, online identification of the inertia parameters and the Center of Mass (CoM) position of the payload for the quadruped robots draw an increasing interest. This study presents an adaptive controller based on the online payload identification for the high payload capacity (the ratio between payload and robot's self-weight) quadruped locomotion. We name it as Adaptive Controller for Quadruped Locomotion (ACQL), which consists of a recursive update law and a control law. ACQL estimates the external forces and torques induced by the payload online. The estimation is incorporated in inverse-dynamics-based Quadratic Programming (QP) to realize a trotting gait. As such, the tracking accuracy of the robot's CoM and orientation trajectories are improved. The proposed method, ACQL, is verified in a real quadruped robot platform. Experiments prove the estimation efficacy for the payload weighing from 20 $kg$ to 75 $kg$ and loaded at different locations of the robot's torso.
\end{abstract}

\section{Introduction}
\IEEEPARstart{L}{egged} robots exhibit remarkable maneuvering capability of traversing rough terrains \cite{lee2020learning, kim2020dynamic, frontierluo2021}. This capability enables the legged robots to have great potential for transportation with heavy payload in daily life. Designing and controlling such machines has motivated considerable research, and several quadruped platforms have been designed and demonstrated superior topography adaptability \cite{bledt2020extracting}.

\begin{figure}
    \centering
    \setlength{\abovecaptionskip}{0.1 cm}
    \setlength{\belowcaptionskip}{-30 cm}
    \includegraphics[width=2.2 in]{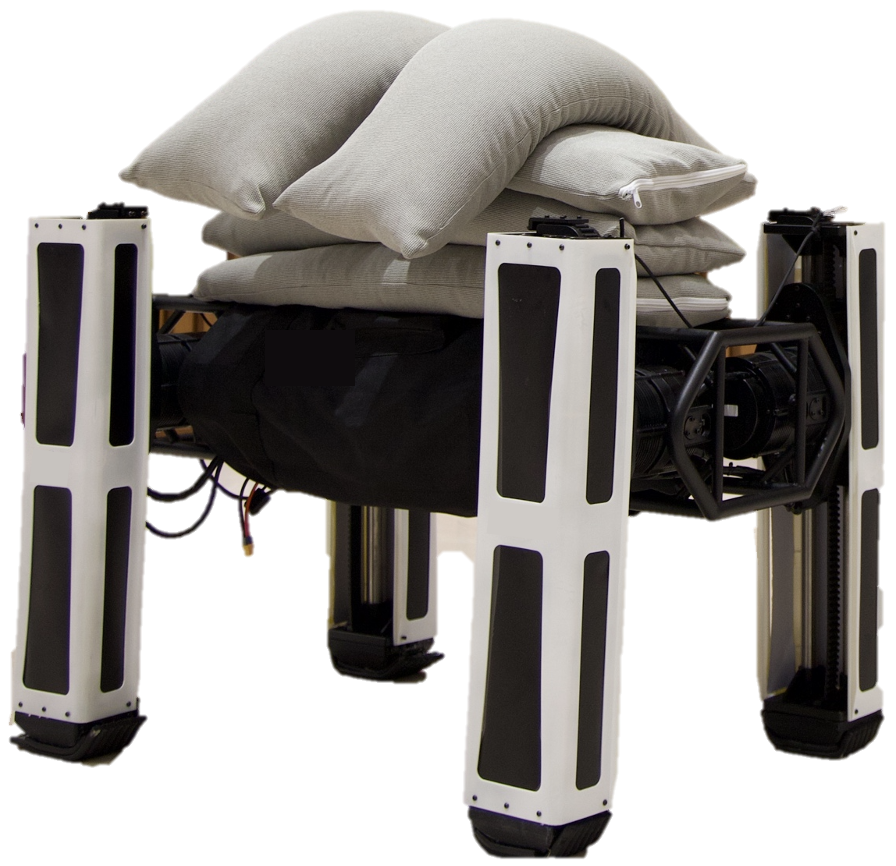}
    \caption{The quadruped robot with high payload capacity, Kirin, is used for the study of the unknown payload identification and adaptive control. It is built with prismatic legs, aiming for capacity of heavy payload carrying. There are three degrees of freedoms (DoFs) for each leg.}
    \label{fig:cover}
\end{figure}

Maximizing the legged robot's maneuverability is one of the most attractive research topics in the locomotion community. The well-known MIT Cheetah robot incorporates virtual leg compliance into its controller and realizes 2D running on the treadmill up to 6 $m/s$ with trot gait in the Sagittal plane \cite{hyun2014high}. StarlETH is another representative quadruped robot developed by ETH Zurich, on which the hierarchical Operational Space Control \cite{hutter2014quadrupedal} is adopted to separate dynamical constraints from trajectory tracking tasks. This enables StarlETH to trot over rough terrains with loose and slippery obstacles\cite{6930798}. Along this line of research, a Whole-Body-Control method  is integrated in the locomotion controller on ETH ANYmal, which significantly improves the robot's agility \cite{7989557}. MIT mini Cheetah, a lightweight version of the MIT Cheetah series, is capable of highly dynamic gaits including 3D trotting and galloping up to 2.5 $m/s$ through convex model predictive control \cite{bledt2019implementing}. Unitree A1 is another lightweight quadruped robot that has performed dynamic locomotion over rough terrains \cite{kumar2021rma}. Besides agile quadruped locomotion, equipping the robot with a payload-carrying ability is also a critical research topic. Bigdog, the first field legged robot that leaves the lab, is hydraulically actuated with impressive robustness to the external disturbances. This robot weighs about 109 $kg$ and can carry a payload weighing more than 150 $kg$ \cite{RAIBERT200810822}. HyQ is another hydraulic robot developed by IIT \cite{hyq2011}. It weights about 80 $kg$ and the peak torque of its hydraulic joint is around 180 $Nm$ which provides the robot extremely large payload-carrying ability \cite{7587429}. Different from traditional rotation joint, baby elephant developed by Shanghai Jiaotong University has a parallel-leg. This electric-hydraulic driven robot weights 130 $kg$ and can carry payload up to 100 $kg$ \cite{zhang2014trot}. Due to the technology limitations, including the electric motor constraints, it is yet extensively explored in the area of the payload-carrying on the electrically-actuated quadruped robots.

Traditional control theories are effectively deployed on quadruped robots, which demonstrated great robustness and agility \cite{kim2020dynamic, bledt2019implementing}. To date, most of these advanced locomotion controllers require accurate robot models, and the predicted control torques to drive the robot heavily rely on the accuracy of robot models, including the link inertia and CoM positions \cite{hutter2014quadrupedal, luo3dwalking, luo2019robust}. However, in the real world, the robot may be commanded to carry unknown payload for transporting goods. In such cases, the control methods that highly depend on deterministic robot models are prone to failures. Motivated by this problem, this paper explores an adaptive controller based on online payload identification for an electric-actuated quadruped robot to handle the unknown payload. The contributions of this letter lie in the following twofold:
\begin{itemize}
    \item An online payload identification algorithm based on a recursive formulation is devised for a high Payload Capacity (PLC) quadruped robot. This algorithm guarantees the fast convergence of the identification.
    \item An adaptive controller based on the online payload identification is verified for PLC from $0.2$ to $1.5$. To our best knowledge, it is the first time to deploy the adaptive control for a wide PLC range on an electrically actuated quadruped robot.
\end{itemize}

The rest of the letter is organized as follows. The related work is reviewed in Section II. Section III introduces the model and dynamics control of the quadruped robot. The payload identification and adaptive control are proposed in Section IV. Section V shows the experimental results. This line of research is concluded in Section VI.

\section{Related Work}

Over the last few decades, research about the robot parameters identification and adaptive control for robots with payload has achieved evident improvements. The inertial parameters identification for legged locomotion has been a critical research topic \cite{5379531, luo2020estimation}. These identification methods are mostly designed offline since the robot model parameters are usually deterministic. These methods have been verified in several robots such as UT-$\mu$2 \cite{1545124}, a small-size humanoid robot, and the quadruped robot HyQ \cite{8678400}. However, for the robot's torso, these methods play a limited role in identification due to the unknown payload. By far, there are two main approaches to solve this issue. One is to identify the torso's parameters online, taking the robot's torso and the payload as a rigid body. Research such as \cite{slotine1987adaptive} proposes an online inertial parameters identification for a manipulator. However, it is only appropriate for the fixed-base rigid body system. HyQ overcomes the shortcomings of this method and proposes a combination of techniques that guarantee the robot locomotion stability \cite{8206367}. These methods have been verified on HyQ in a static walking gait. Scalf-III, a hydraulic actuated heavy-duty quadruped robot developed by Shandong University, proposes a CoM estimation and adaptation method in dynamic trot gait and verifies it in simulation \cite{9294027}. Another approach is to incorporate the adaptive control into the traditional quadruped locomotion controller. A $L_1$ adaptive control theory is proposed for legged robots and verified in simulation \cite{nguyen20151, sombolestan2020adaptive}. 

To our best knowledge of the existing work, the adaptive control with online unknown payload identification is yet fully explored for high capacity quadruped locomotion. In this study, an adaptive control with high-payload online identification is proposed for an electrically-actuated quadruped robot. The robot is named Kirin as shown in Fig. \ref{fig:cover}. The leg mechanism is designed to be prismatic so as to greatly increase the payload capacity. Experiments are conducted on Kirin to verify the effectiveness of ACQL.

\section{Dynamics Control for Quadruped Locomotion}
The legged locomotion model and control relate to the effects of the forces and the torques exerted on the robot, which originate from the robot and payload's gravity, and the physical interaction with the environment. In this study, the desired forces and torques are computed by:

\begin{equation}
\begin{cases}
\begin{aligned}
    F_b & = K_p^f(r^d_b-r^a_b)+K_i^f\int{(r^d_b-r^a_b)}+\\
     & \quad K_d^f(v^d_b-v^a_b)+ma_b+\sum{m_ig}+m_pg
\end{aligned} \\ \\
\begin{aligned}
    T_b & = K_p^t\log{(q^d_b \cdot (q^a_b)^{-1})} \\
    & + K_i^t\int{\log{(q^d_b \cdot (q^a_b)^{-1})}} \\
    & + K_d^t\log{(\omega^d_b \cdot (\omega^a_b)^{-1})} \\
    & + \sum{r_i\times m_ig}+r_p\times m_pg
\end{aligned}
\end{cases},
\label{eq:model}
\end{equation}
where $r^d_b$, $v^d_b$, $a_b$, $q^d_b$, $\omega^d_b$ represent the desired position, linear velocity, linear acceleration, rotation matrix and angular velocity of the robot torso respectively. $r^a_b$, $v^a_b$, $q^a_b$ and $\omega^a_b$ represent the actual position, linear acceleration, rotation matrix and angular velocity of the robot torso respectively. $m_ig$ represents the gravitational force acting on the body $i$. $r_i$ represents the corresponding position vector from the origin of the inertial frame to the CoM of the body $i$. $m$ represents the total mass of the robot, which is equal to the summation of $m_i$. $m_pg$ is the gravity of the payload and $r_p$ means the corresponding position vector of $m_pg$. Different from $m_ig$ and $r_i$ that can be generated from computer-aid design software directly, $m_pg$ and $r_p$ are the unknown, dynamically changing parameters that can not be ignored. $K^{(\cdot)}_{(\cdot)}$ represents the diagonal gain matrices which need to be manually tuned in the experiment. For a matrix $M \in$ SO(3), the logarithm operation is:
\begin{equation}
\begin{aligned}
\log(M)=
\begin{cases}
    \frac{1}{2}(M-M^T)    \qquad  \qquad  & d\to1 \\
    \arccos{d} (M-M^T)/(2\sqrt{1-d^2})    & other \\
\end{cases}
\end{aligned},
\end{equation}
where $d = (trace(M)-1)/2$. 

The first three components for each equation's right side in (\ref{eq:model}) are the PID tracking controllers and the remaining components represent the feedforward controller. 

In this study, an inverse-dynamics-based Quadratic Programming is adopted to realize the quadruped locomotion, including the trotting gait. The inverse dynamics solver outputs the desired force $F_b$ and torque $T_b$ exerted on the torso of the robot to track the desired motion. (\ref{eq:model}) is formulated as a Quadratic Programming (QP) problem, by which the contact force $F^{d}$ is computed. The QP formulation is given as:

\begin{equation}
\begin{aligned}
    F^{d} & = \mathop{\arg\min}_{F} (AF-B)^T Q (AF-B) + F^T R F \\
    & s.t. 
    \begin{cases} 
    \quad \tau_{min} \leq J^{-1} F \leq \tau_{max} \\
    -\mu F_{iz} \leq  F_{ix} \leq \mu F_{iz} \\
    -\mu F_{iz} \leq  F_{iy} \leq \mu F_{iz} \\
    \quad D_iF_i = 0 \\
    \end{cases}
\end{aligned},
\label{eq:3}
\end{equation}
where $\tau_{min}$ and $\tau_{max}$ are the robot's minimal and maximum joint torque. $F_{ix}, F_{iy}$, and $F_{iz}$ are the components of each foot's contact force vector. $\mu$ is the coefficient of friction between the contact foot and the ground. $D_i$ represents the matrix which selects the feet that dose not contact the ground. $\hat{r}_{bi}$ is the skew-symmetric matrix defining the cross product of the position vector $r_{bci}$ from base to contact foot. $F^d \in \Re^{3N \times 1}$ is the concatenated vector of the contact forces. $N$ represents the number of legs that contact the ground. The diagonal matrix $Q$ and $R$ are the weight matrix which needs to be adjusted via experiment as well. $A$ and $B$ are given as:
\begin{equation}
\begin{cases}
    A= \begin{bmatrix}
    \cdots & I & \cdots\\
    \cdots & \hat{r_{bi}} & \cdots\\
  \end{bmatrix} \\
  \\
  B= \begin{bmatrix}
    F_b\\
    T_b\\
  \end{bmatrix}
\end{cases}.
\label{eq:4}
\end{equation}

From (\ref{eq:model})-(\ref{eq:4}), The locomotion controller requires accurate model parameters, especially the unknown payload parameters. The identification of the payload will be introduced in detail in section IV.

\section{Payload Identification and Adaptive Control}
The high-payload identification and adaptive control for quadruped locomotion are introduced in this section. When quadruped robots are deployed for goods transportation, the payload is mostly unknown and fluctuating. The variable payload incurs significant disturbances to the robot balance. In this study, an adaptive control for quadruped locomotion (ACQL) is proposed to tackle this issue. For simplicity of analysis within the scope of this study, ACQL is based on the following assumptions:

(i) The payload effects to the robot are identified as a force and a torque. Although the payload is always in surface contact with the robot, only the force and moment acting on the robot are considered in the robot's dynamics model. Therefore, the specific force distribution is not taken into account.

(ii) The force and the torque to be identified do not change during locomotion. The identification method proposed in this study mainly relies on the robot's orientation error. During the dynamic gaits, orientation errors easily bring noises in the identification. Therefore, it is preferred to estimate these parameters in a static pose, such as standing on the ground.

\begin{figure}
    \centering
    \setlength{\abovecaptionskip}{-0.3 cm}
    \setlength{\belowcaptionskip}{-10 cm}
    \includegraphics[width=3.45 in]{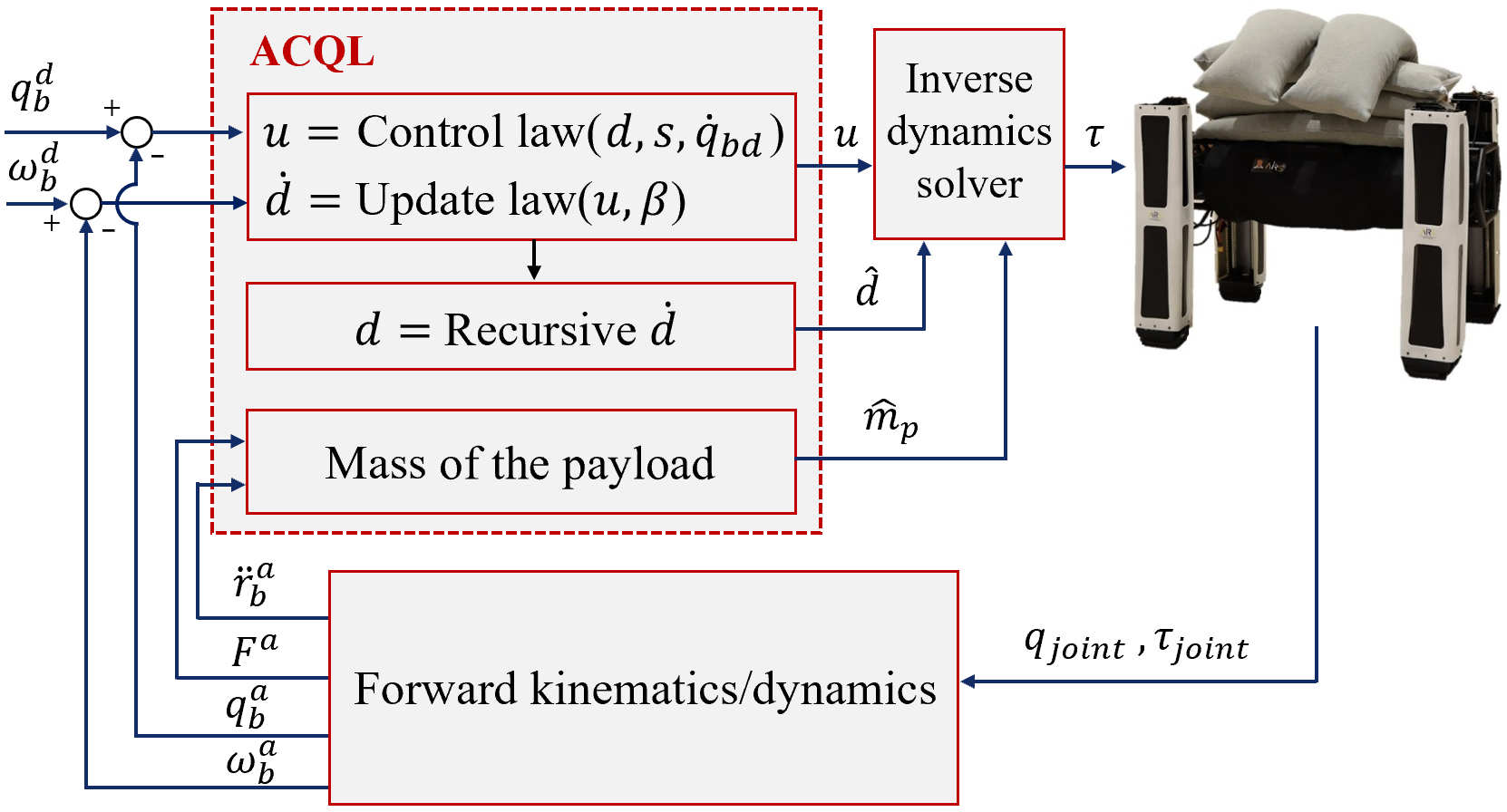}
    \caption{Control scheme of the adaptive control with online unknown payload identification. The Adaptive Control for Quadruped Locomotion (ACQL) identifies the mass of the payload directly and generates an update law. The recursive result of the update law is the moment of the payload concerning the robot. ACQL also generates a control law by which the robot can adjust its posture. The torques of joints are calculated via inverse dynamics based on Quadratic Programming (QP) solver.}
    \label{fig:adaptive}
\end{figure}

Based on the assumptions, this section introduces the quadruped dynamics model, ACQL and demonstrates the stability proof. ACQL is based on the quadruped dynamics model and consists of update law and control law.

\subsection{Quadruped Robot Model}

The dynamic model of a quadruped robot is given by:
\begin{equation}
\begin{cases}
\begin{aligned}
     m\ddot{r}^a_b &=\sum F_i-\sum{m_i}g-m_pg \qquad \qquad \\
    \frac{d}{dt}(I\omega^a_b) &=\sum r_{bci}\times F_i+r_p\times m_pg \\
    & +\sum r_i\times {m_i}g
\end{aligned}
\end{cases},
\label{eq:5}
\end{equation}
where $I \in \Re^{3 \times 3}$ is the inertial of the robot. $F_i$ represents the contact force of each foot $i$. The left component of the second equation of (\ref{eq:5}) can be extended as:
\begin{equation}
     \frac{d}{dt}(I\omega^a_b) = I\dot{\omega}^a_b+\omega^a_b\times (I \omega^a_b).
\label{eq:6}
\end{equation}

With the aforementioned assumption (ii), when the robot stands on the ground with all four legs, the robot's angular velocity is small, and the precession and nutation of $\omega^a_b\times (I \omega^a_b)$ of  (\ref{eq:6}) contribute little to the dynamics of the robot. Thus, this component can be discarded, and the second equation of (\ref{eq:5}) can be changed to be:
\begin{equation}
    I\dot{\omega}^a_b=\sum r_{bci}\times F_i+r_p\times m_pg+\sum r_i\times {m_i}g.
    \label{eq:8}
\end{equation}

\subsection{Identification of Payload Mass and Moment}

The scheme of ACQL for the identification of the payload's mass and moment is shown in Fig. \ref{fig:adaptive}. In the proposed ACQL, the payload's estimated mass is given by an analytical solution which is computationally efficient. A control law is proposed to adjust the robot's posture while an update law is used to identify the moment of the payload concurrently.

The first equation of (\ref{eq:5}) can be rewritten and the estimated mass of the payload $\hat{m}_p$ is given by:

\begin{align}
    \hat{m}_p=\frac{1}{g}\sum F_i-\sum{m_i}-\frac{m\ddot{r}^a_b}{g}.
\end{align}

To estimate the moment of the payload, (\ref{eq:8}) can be rewritten with some substitution and the robot dynamics can be formulated in a standard state-space form:
\begin{equation}
\begin{cases}
\begin{aligned}
    \dot{x}_1 & =x_2\\
    \dot{x}_2 & =Bu+d+k\\
    \dot{\hat{d}} & = w
    \label{eq:9}
\end{aligned}
\end{cases},
\end{equation}
where $x_1$ represents the robot orientation. $x_2$ represents the angular velocity of the robot. $B$ represents $I^{-1}$. $u$ represents $\sum r_{bci}\times F_i$. $k$ represents $\sum{r_i\times {m_i}g}$. $d$ represents $I^{-1}r_p\times m_pg$ which is the parameter to be identified. $\hat{d}$ represents the estimated value of $d$. $w$ represents the update law $\dot{\hat{d}}$.

In the quadruped robot system, there are multiple variables that can be used as the observed quantity such as orientation, angular velocity of torso and rotation matrix. In this study, the robot orientation is selected as the tracking object. The tracking error can be written as $\widetilde{x}=x_1-x_{1d}$ and the first and second derivative of the tracking error are $\dot{\widetilde{x}}=\dot{x_1}=x_2$ and $\ddot{\widetilde{x}}=\dot{x_2}$ respectively. Consider the function:

\begin{equation}
   s=\dot{\widetilde{x}}+\lambda \widetilde{x},
   \label{eq:11}
\end{equation}
where $\lambda$ is a positive definite matrix. Combined with (\ref{eq:9}), the derivative of $s$ is given by:

\begin{equation}
\begin{aligned}
   \dot{s} & =\ddot{\widetilde{x}}+\lambda \dot{\widetilde{x}}\\
   & = \dot{x}_2+\lambda x_2\\
   & = Bu+d+k+\lambda x_2.
   \label{eq:12}
\end{aligned}
\end{equation}

Based on the derivation, the control law is devised as:
\begin{equation}
    u=-B^{-1}(\hat{d}+k+cs+\lambda x_2),\\
    \label{eq:13}
\end{equation}
where $c$ is a positive definite matrix. Put (\ref{eq:13}) in (\ref{eq:12}) we have:
\begin{equation}
   \dot{s}=-cs+ \widetilde{d},
\label{eq:14}
\end{equation}
where $\widetilde{d}=d-\hat{d}$. From (\ref{eq:14}), it can be concluded that $\dot{s}$ will asymptotically converge to zero if $\widetilde{d}$ asymptotically converges to zero, and in this case $s=0$. The proof of this conclusion will be demonstrated in the next subsection. 

The update law $\dot{\hat{d}}$ is devised based on the control law and the aforementioned assumptions. Consider the manifold:
\begin{equation}
    M=\{(x_1,x_2,d) \in \Re^{3 \times 3}\mid\hat{d}-d+\beta(x_1,x_2)=0\},
\end{equation}
where $\beta(x_1,x_2)$ is the estimation error function to be designed. As such, the problem is reformulated into the design of an appropriate function $\beta(x_1,x_2)$ to ensure the manifold to be invariant and attractive. Define the manifold as:
\begin{equation}
   z=\hat{d}-d+\beta(x_1,x_2).
\end{equation}
\begin{algorithm}[t]
\caption{ACQL algorithm} 
\label{alg_adaptive_control}  
\renewcommand{\algorithmicrequire}{\textbf{Input:}}
\renewcommand{\algorithmicensure}{\textbf{Output:}}
\begin{algorithmic}[1]
{\fontsize{9pt}{12.5pt}\selectfont
\REQUIRE $ q^d_b,\omega^d_b,r^d_b,v^d_b,m_i,r_i,q^a_b,\omega^a_b,q_{joint},\dot{q}_{joint},\tau_{joint}$
\ENSURE $F^d$
\STATE $ ^Br_{i},^Bv_{i},^B\ddot{r}_{i} \Leftarrow$ Forward kinematics $(q_{joint},\dot{q}_{joint}) $
\STATE $ ^Br^a_b,^Bv^a_b,^B\ddot{r}^a_b \Leftarrow ^Br_{i},^Bv_{i},^B\ddot{r}_{i} $
\STATE $ R \Leftarrow q^d_b $
\STATE $ r^a_b,v^a_b,\ddot{r}^a_b \Leftarrow R^Br^a_b,R^Bv^a_b,R^B\ddot{r}^a_b $
\STATE $F_{a}\Leftarrow \tau_{joint}$ 
\STATE $ \hat{m}_p \Leftarrow m_i, \ddot{r}^a_b,F_{a}$
\STATE $ e_{qb} \Leftarrow \log{(q^d_b\cdot(q^a_b)^{-1})}$
\WHILE {$ e_{qb} > e_{threshold}$}
\STATE $ u \Leftarrow -B^{-1}(\hat{d}+k+cs+\lambda x_2)$ (control law)
\STATE $ \dot{d} \Leftarrow -M(Bu+\hat{d}+\beta(x_1,x_2)+k)$ (update law) 
\STATE $ \hat{d}_{k+1} = \dot{d}\Delta t + \hat{d}_k$
\STATE $ T_b \Leftarrow Iu$
\STATE $ F_b \Leftarrow m_i,\hat{m}_P,\ddot{r}^a_b$
\ENDWHILE
\STATE $ e_{qr} \Leftarrow r^a_b,r^d_b$
\STATE $ \dot{e}_{qr} \Leftarrow v^a_b,v^d_b$
\STATE $ \dot{e}_{qb} \Leftarrow \log{(\omega^d_b\cdot(\omega^a_b)^{-1})}$
\STATE $ F_b \Leftarrow e_{qr},\dot{e}_{qr},m_i,\hat{m}_p,a_b$
\STATE $ T_b \Leftarrow e_{qb},\dot{e}_{qb},r_i,m_i,r_p,\hat{m}_p$
\STATE $B \Leftarrow F_b, T_b$
\STATE $i \Leftarrow r^a_b, r_i$
\STATE $ F^d \Leftarrow \mathop{\arg\min}\limits_{F} (AF-B)^T Q (AF-B) + F^T R F $
} \\
\end{algorithmic}  
\end{algorithm}
The derivative of $z$ is:
\begin{equation}
   \dot{z}=\dot{\hat{d}}+\dot{\beta}(x_1,x_2).
   \label{eq:16}
\end{equation}

Substitute the robot dynamic system (\ref{eq:9}) into (\ref{eq:16}):
\begin{equation}
\begin{aligned}
   \dot{z} & =w+\frac{\partial \beta}{\partial x_1}x_2+\frac{\partial \beta}{\partial x_2}\dot{x}_2\\
   & = w+\frac{\partial \beta}{\partial x_1}x_2+\frac{\partial \beta}{\partial x_2}(Bu+d+k)\\
   & = w+\frac{\partial \beta}{\partial x_1}x_2+\frac{\partial \beta}{\partial x_2}(Bu+\hat{d}-z+\beta(x_1,x_2)+k).\\
\end{aligned}
\end{equation}

The update law $\dot{\hat{d}}$ and manifold $z$ can thus be designed as:
\begin{equation}
\begin{cases}
   \dot{\hat{d}}= -\frac{\partial \beta}{\partial x_1}x_2-\frac{\partial \beta}{\partial x_2}(Bu+\hat{d}+\beta(x_1,x_2)+k)\\
   \dot{z} = -\frac{\partial \beta}{\partial x_2}z
\end{cases}.   
\end{equation}

In order to make sure the system $z$ is Lyapunov stable and reduce the computation complexity, the estimation error function $\beta(x_1,x_2)$ is designed as:
\begin{equation}
   \beta(x_1,x_2) = \begin{bmatrix}
    k_1\omega^a_{bx}\\
    k_2\omega^a_{by}\\
    k_3\omega^a_{bz}\\ 
  \end{bmatrix},
\end{equation}
where $k_1,k_2,k_3$ are all greater then zero. $\omega^a_{bx},\omega^a_{by},\omega^a_{bz}$ are the components of vector $\omega^a_b$. 

The update law $\dot{\hat{d}}$ and the function $z$ can thus be written as:

\begin{equation}
\begin{cases}
\begin{aligned}
   \dot{\hat{d}} & =-M(Bu+\hat{d}+\beta(x_1,x_2)+k)\\
   \dot{z} & =-Mz
   \label{eq:20}
\end{aligned}
\end{cases},
\end{equation}
where $M$ is given as:
\begin{equation}
\begin{aligned}
      M = \begin{bmatrix}
    k_1 & 0 & 0\\
    0 & k_2 & 0\\
    0 & 0 & k_3\\
  \end{bmatrix}. \\
\end{aligned}
\end{equation}

The ACQL algorithm is shown in Algorithm \ref{alg_adaptive_control}. In terms of Lyapunov's second method for stability, it proves that (\ref{eq:20}) asymptotically converges to zero. The moment of the payload concerning the robot can be identified with (\ref{eq:13}) and (\ref{eq:20}). It is noteworthy that the control law of (\ref{eq:13}) is the aggregate moment of all four foot-end forces. This value is redistributed to each foot with the QP solver. 
\vspace{-5mm}
\subsection{Stability Proof of ACQL}

The proof of the convergence of $s$ in the identification of the payload moment section is shown in this subsection. Firstly, it is proven that there exists $s=0$. Substitute (\ref{eq:13}) into (\ref{eq:20}):

\begin{equation}
\begin{aligned}
    \dot{\hat{d}} & =-M(-cs-\lambda x_2+\beta)\\
    & =M[c(\dot{\widetilde{x}}+\lambda \widetilde{x})+\lambda x_2-Mx_2]\\
    & =M[c\lambda(x_1-x_{1d})+(c+\lambda-M)Mx_2].
    \label{eq:22}
\end{aligned}
\end{equation}

It can be summarized by (\ref{eq:22}) that the update law  $\dot{\hat{d}}$ depends on the robot orientation error and its angular velocity. By choosing appropriate positive definite matrices of $c,\lambda$, and $M$, the robot is able to adjust its orientation until its error approaches zero, which implies the function $s$ will eventually approach zero.

Secondly, it is proved that (\ref{eq:14}) asymptotically converge to zero. since $\widetilde{d}$ is designed to asymptotically converge to zero as described above, only $\dot{s}=-cs$ needs to be considered. Define the Lyapunov function as:
\begin{equation}
    V(s)=\frac{1}{2}s^2.
    \label{eq:23}
\end{equation}

It can be seen in (\ref{eq:23}) that $V=0$ if and only if $s=0$ and $V>0$ when $s\neq 0$. The derivative of $V$ is:
\begin{equation}
    \dot{V}(s)=s \dot{s}=-cs^2\leq 0.
\end{equation}

According to the second method of Lyapunov, the system as described in (\ref{eq:14}) is proven to asymptotically converges to zero. Therefore, ACQL is proven to be Lyapunov stable.

\section{Experiment}
\subsection{Experiment Platform}

\begin{figure}
    \centering
    \includegraphics[width=2.6 in]{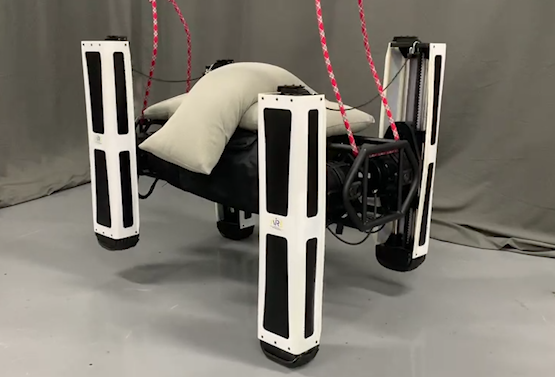}
    \caption{The quadruped robot for high payload capacity locomotion. It is an electrically-actuated quadruped robot with 12 degrees-of-freedom (DoFs) and is named as Kirin. The knee joint is designed to be prismatic to enhance the payload-carrying capacity in dynamic locomotion.}
    \label{fig:robot_dof}
\end{figure}

The experiments to verify the verification of the effectiveness of the proposed method are conducted on the quadruped robot, Kirin. The Kirin is an electrically actuated quadruped robot developed for high payload capacity. Kirin has one hip roll joint, one hip pitch joint, and one knee joint on each leg (as shown in Fig. \ref{fig:robot_dof}). The total weight of Kirin is around 50 $kg$ and payload capacity can reach up to 2.0 (with at least 100 $kg$ payload). The hip roll and pitch joints are revolute joints, while the knee joint is designed as the prismatic joint. The controller of Kirin consists of Nvidia TX2, on which RT-Linux is installed. The high-level forward and inverse dynamics, online payload identification, and adaptive control algorithms are running on RT-Linux. The size of Kirin is 700 $mm$ $\times$ 240 $mm$ $\times$ 600 $mm$. The peak joint torque and velocity are around 200 $Nm$ and 150 $rpm$ respectively. The actuator specifications of all the joints are uniform due to the commercial constraints. The joint power and the mechanism design are capable of supporting the heavy payload carrying and dynamic gait. The specifications of the quadruped robot Kirin are listed in Table. \ref{table:1}.

\begin{table}
  \caption{Specifications of the Quadruped Robot Kirin}
  \centering 
  \small
  \begin{threeparttable}
  \begin{tabular}{p{4.4cm} | p{3.2cm} }
  \hline \hline
  \small
  Property & Parameters \\
  \hline
  Dimensions (L$\times$W$\times$H) ($mm$) & 700 × 240 × 600 \\
  \hline
  Active DoF number & 12 \\
  \hline
  Total weight ($kg$) & 50 \\
  \hline
  Maximum payload \tnote{*} (kg) & 100+ \\
  \hline
  Motion range of hip roll joint ($^{\circ}$) & 280 \\
  \hline
  Motion range of hip pitch ($^{\circ}$) & 360 \\
  \hline
  Motion range of knee joint ($mm$) & 300 \\
  \hline
  Electric actuator & Customized QDD \\
  \hline
  Computing board & Nvidia Jetson TX2 \\
  \hline
  Joint peak torque (Nm) & 200 \\
  \hline
  Joint peak speed (rpm) & 150 \\
  \hline
  Motor driver & G-SOLWH120/100EES \\
  \hline
  \end{tabular}
  \begin{tablenotes}
    \footnotesize
    \item[*] Due to the safety consideration, 100kg is the experiment result at the current stage. More payload is to be tested.
  \end{tablenotes}
  \end{threeparttable}
  \label{table:1}
\end{table}

\subsection{ACQL Test with Varied Payload}

\begin{figure}
    \centering
    \setlength{\abovecaptionskip}{-0.5 cm}
    \setlength{\belowcaptionskip}{-20 cm}
    \includegraphics[width=3.45 in]{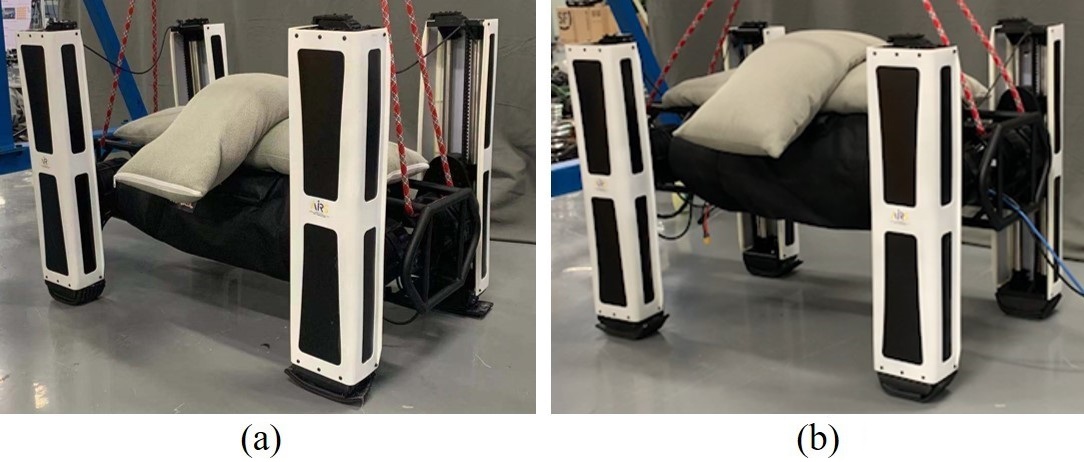}
    \caption{The experiment scenario of the payload mass identification. (a) is the initial posture of the robot. (b) is the posture of the robot when the identification ends.}
    \label{fig:payload_init}
\end{figure}

\begin{figure}[ht]
    \centering
    \setlength{\abovecaptionskip}{-0.1 cm}
    \setlength{\belowcaptionskip}{-20 cm}
    \includegraphics[width=2.8 in]{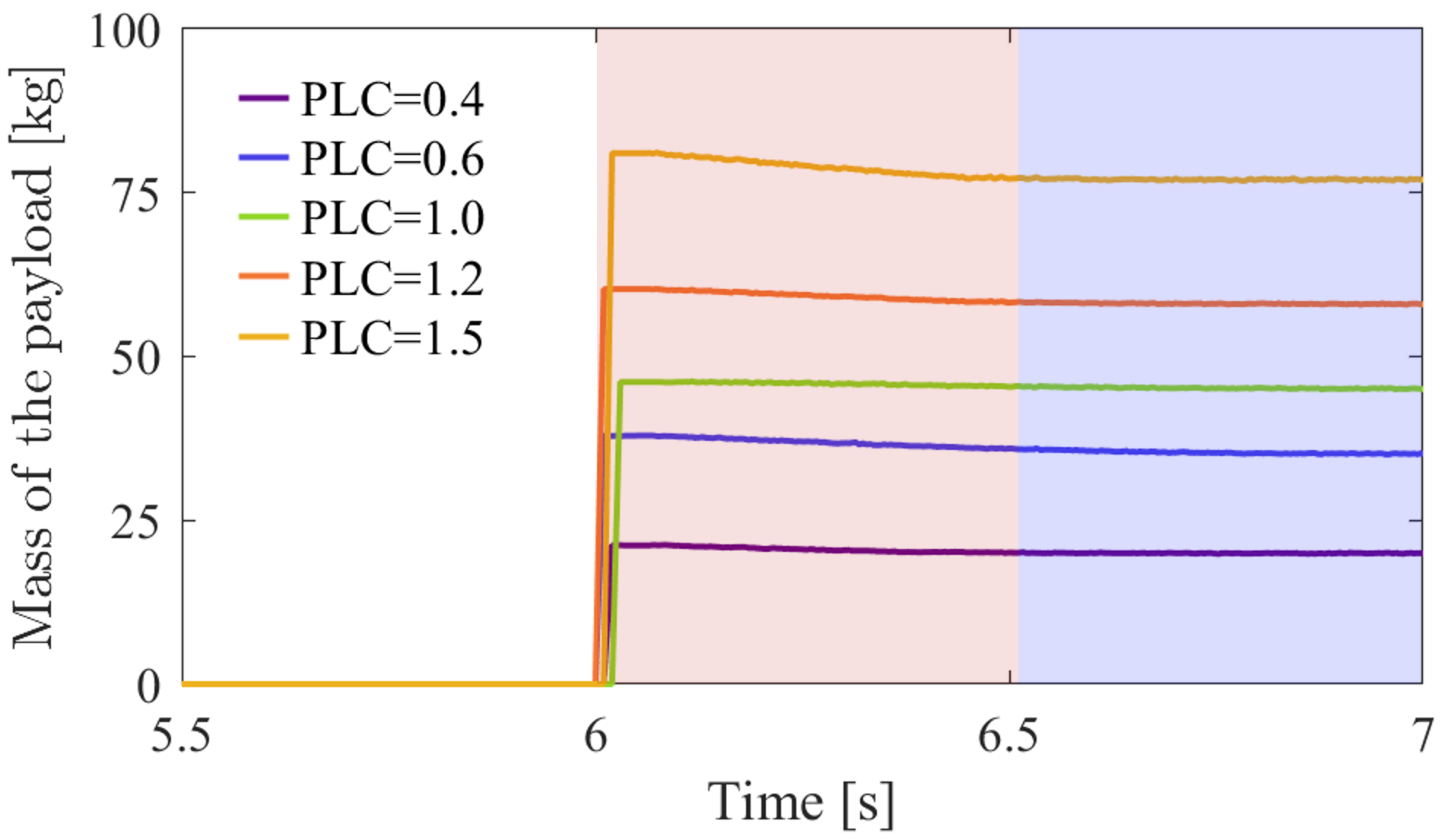}
    \caption{The mass of the payload identified in this experiment. Payload varies from 20$kg$ to 75$kg$. Payload varies from 20$kg$ to 75$kg$, which corresponds to PLC (Payload Capacity) from 0.4 to 1.5. The identification starts from 6 $s$ (red region) and ends at around 6.5 $s$ (blue region)}
    \label{fig:mass_id}
\end{figure}

\begin{figure}[ht]
    \centering
    \setlength{\abovecaptionskip}{-0.1 cm}
    \setlength{\belowcaptionskip}{-15 cm}
    \includegraphics[width=3.4 in]{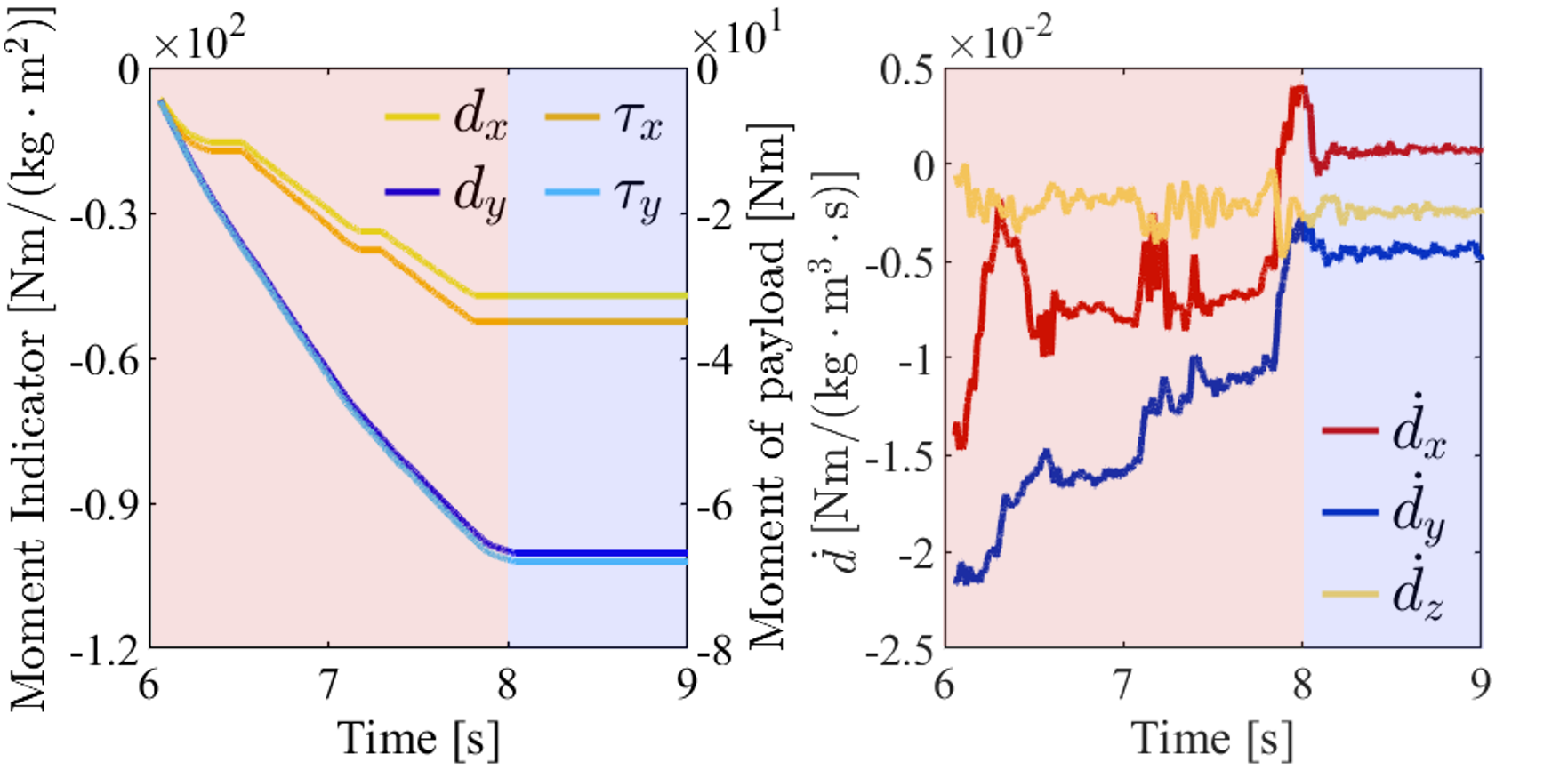}
    \caption{The estimated $d$ and the estimated moment of payload. Since the inertial matrix of the robot is a diagonal matrix, the estimated $d$ and the estimated moment of payload have the same shape with the different values. The recursive process undergoes in the red region and converges in the blue region.}
    \label{fig:d}
\end{figure}

\begin{figure}[!ht]
    \centering
    \setlength{\abovecaptionskip}{-0.5 cm}
    \setlength{\belowcaptionskip}{-6 cm}
    \includegraphics[width=3.45 in]{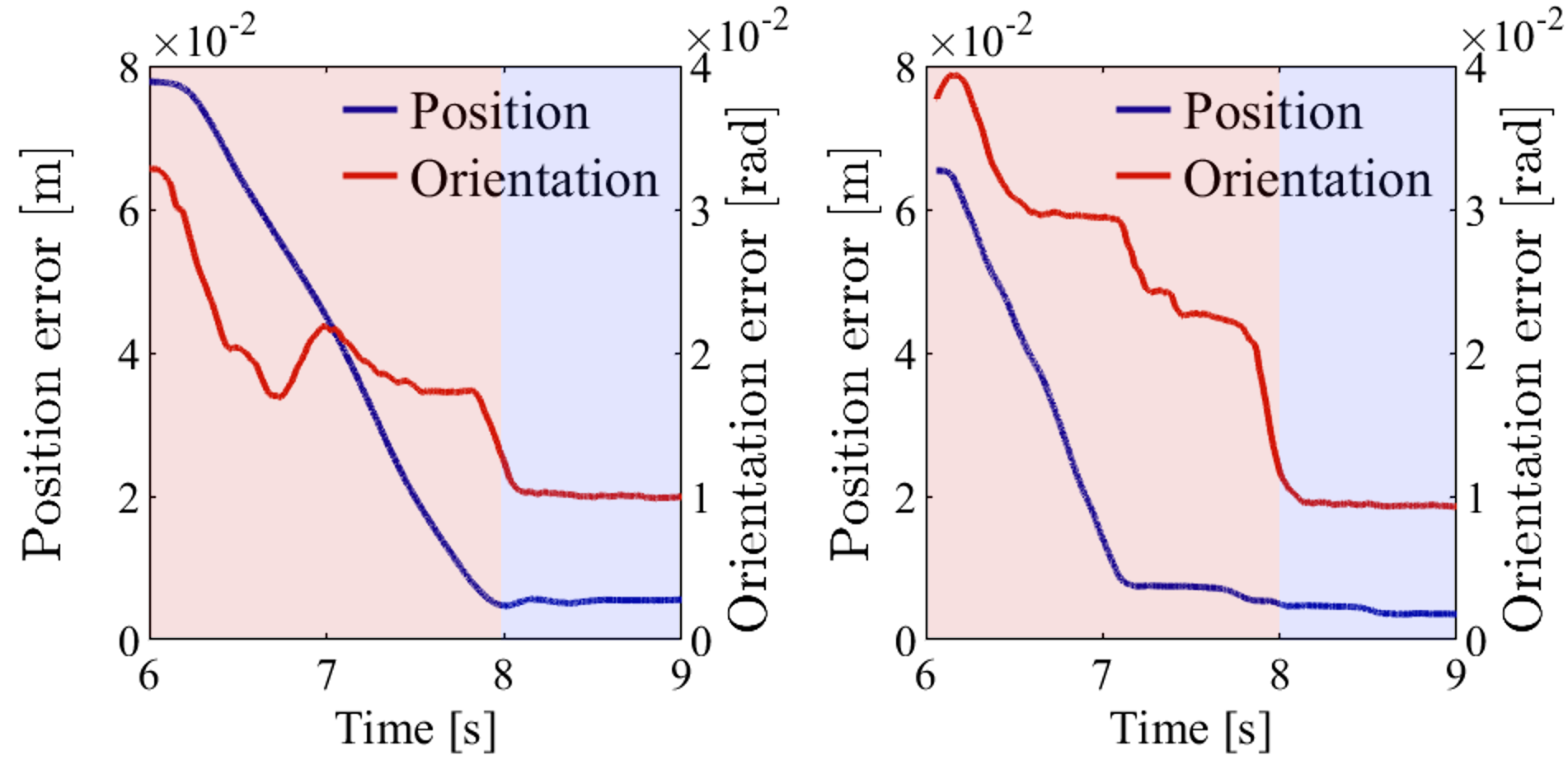}
    \caption{The norm of the robot position error and orientation error. Each corresponds to the payload arrangements in Fig. \ref{fig:sandbag}(a) and (b) respectively. The recursive process undergoes in the red region and converges in the blue region.}
    \label{fig:tracking_error}
\end{figure}

In this section, several experiments were carried out with the Kirin that verifies the efficacy of the proposed ACQL algorithm. As shown in Fig. \ref{fig:payload_init}, the robot initially stands on the ground and carries unknown payloads (sandbags). A predefined control torques are applied to drive the robot to reach the desired height with four legs on the ground supporting the main body. Influenced by the unknown payloads, there are both significant errors in the position and orientation of the robot's torso. Then, the robot will not stop the identification of the mass and the moment concerning the robot generated by the payload and adjusting its posture using the adaptive control proposed above until the errors of the robot's orientation reduce to the predefined thresholds. In the experiments, the thresholds for the orientation convergence are set to 0.01 $rad$ in each rotational direction. The parameters $c, \lambda$, and $M$ in ACQL are set to $0.7 I, 0.7 I$, and $1.3 I$ respectively. These parameters govern the recursive rate of the payload moment identification.

The results of the payload mass identification are shown in Fig. \ref{fig:mass_id}. In the experiment, payload varies from 20 $kg$ to 75 $kg$, which implies that PLC varies from 0.4 to 1.5. Benefiting from our previous work \cite{jin2019joint} and ACQL, the maximum estimated error of the payload mass is about 3 $kg$, which accounts for only 6$\%$ of the robot's mass. The error has little influence on the robot dynamic locomotion. 

\begin{figure}
    \centering
    \setlength{\abovecaptionskip}{-0.1 cm}
    \setlength{\belowcaptionskip}{-5 cm}
    \includegraphics[width=2.6 in]{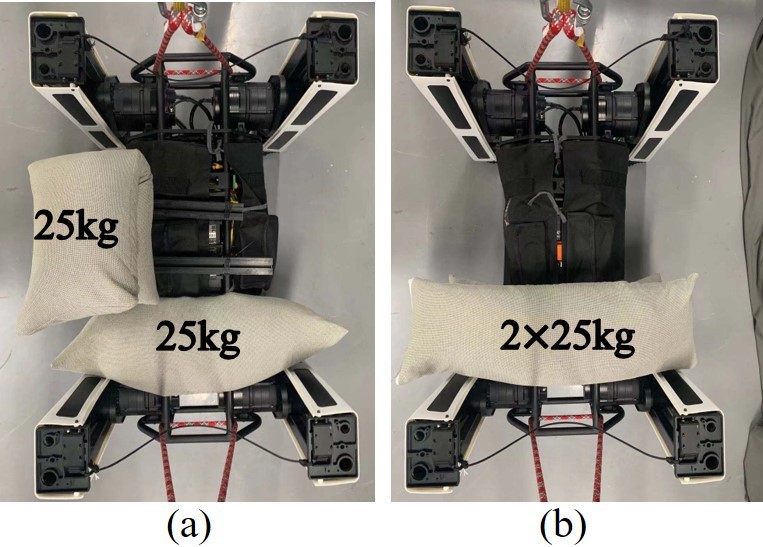}
    \caption{The top view of the quadruped robot in the experiment of identifying the moment with respect to the robot induced by payload. The payload is two sandbags, each of which weighs about 25 $kg$. They are placed on the different location on the robot's back.}
    \label{fig:sandbag}
\end{figure}

\begin{figure}
    \centering
    \includegraphics[width=3 in]{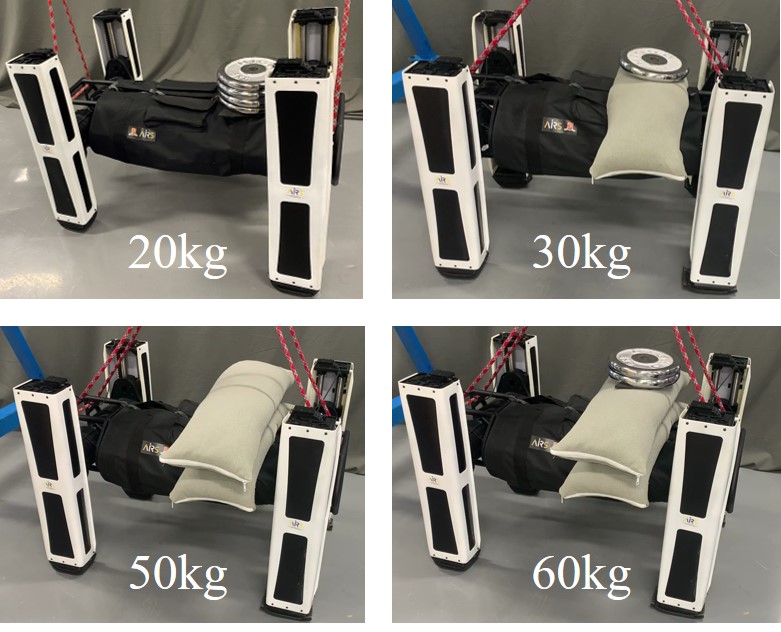}
    \caption{The experiment scenario of different PLC. Payload of different mass are placed in the front on the robot back. Each sandbag weights 25$kg$ while each dumbbell weights 5$kg$.}
    \label{fig:payload_4_weights}
\end{figure}

\begin{figure*}[ht]
    \centering
    \includegraphics[width=7.0 in]{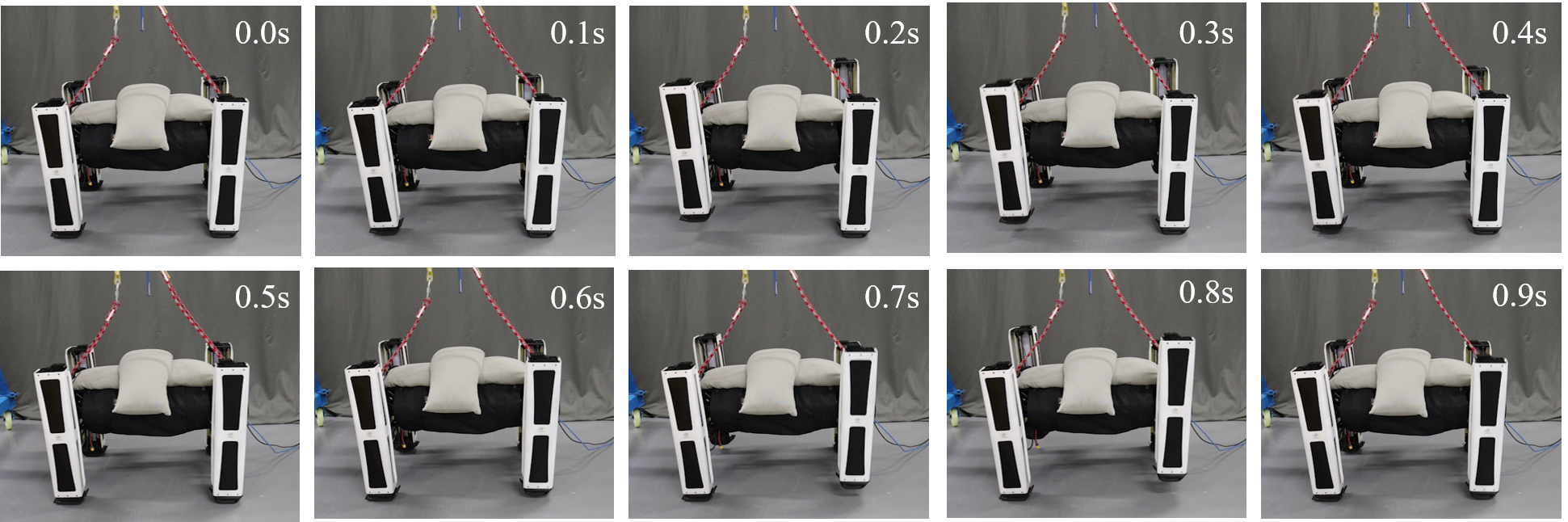}
    \caption{The experiment screenshots of Kirin trotting with payload weighing around 50$kg$. Two sandbags are placed on the back of Kirin, each of which weighs around 25$kg$.}
    \label{fig:exp_screenshot_plc}
\end{figure*}

\begin{figure}[ht]
    \centering
    \setlength{\abovecaptionskip}{-0.4 cm}
    \setlength{\belowcaptionskip}{-5 cm}
    \includegraphics[width=3.45 in]{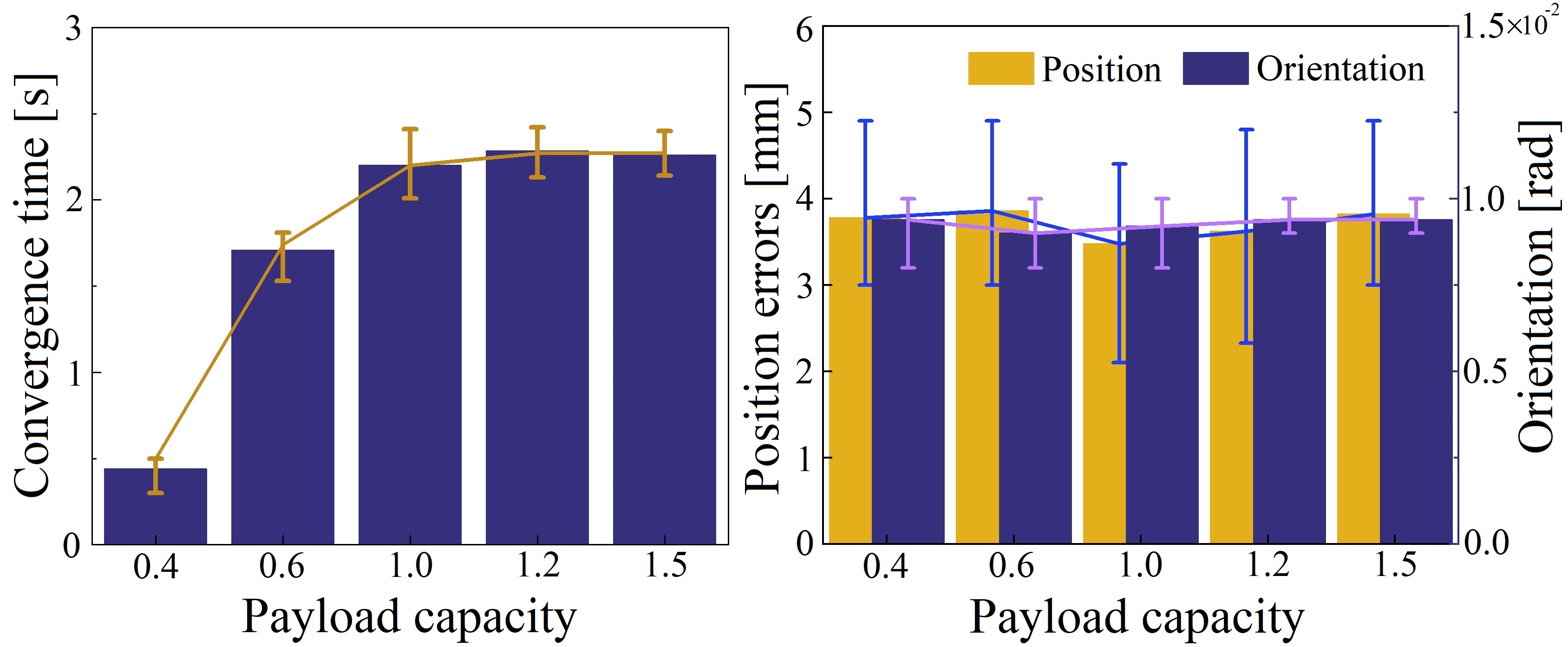}
    \caption{The convergence time and the RMSE of the tracking errors for a series of PLCs with ACQL. ACQL is verified to converge within a wide range of payload with PLC up to 1.5.}
    \label{fig:converge_plc}
\end{figure}

The results of the moment concerning the robot generated by the payload are depicted as shown in Fig. \ref{fig:d} to Fig. \ref{fig:tracking_error}. As shown in Fig. \ref{fig:sandbag} (a), two sandbags (each weighs around 25 $kg$) are loaded on the back of Kirin. These two sandbags are placed in the front and left part on the back of Kirin, away from the robot's original CoM, to verify the effectiveness of ACQL. The result of the derivative of $d$ is shown in Fig. \ref{fig:d}. These values converge to less than 0.005 $Nm/(kg\cdot m^3\cdot s)$ within 2 $s$ when the orientation error reduces below the threshold. The result of the estimated $d$ and the moment of the payload are shown in Fig. \ref{fig:d}. Since the inertial matrix of the robot, generated from CAD, is a diagonal matrix, the estimated $d$ and the estimated moment of payload have the same structure yet with different values. With the measured mass of the payload and the dimensions of the robot depicted in table \ref{table:1}, it can be proved that the estimated moment of the payload is close to its true value. The norm of the robot position and orientation tracking errors is shown in Fig. \ref{fig:tracking_error}, which are usually used to illustrate the identification performance \cite{8793663}. As shown in Fig. \ref{fig:tracking_error}, after the payload identification, the robot adjusts its body with the position and orientation tracking error at around 3 $mm$ and 0.008 $rad$ respectively. To furtherly verify the effectiveness of ACQL, both two sandbags (50 $kg$) are placed in the front of the robot as shown in Fig. \ref{fig:sandbag} (b). The robot has the same tracking performance as shown in the right subfigure of Fig. \ref{fig:tracking_error}.

The convergence time of the proposed ACQL is tested on Kirin with a wide range of PLC. Payload varies from 20 $kg$ to 75 $kg$ which represents PLC from 0.4 to 1.5 as shown in Fig. \ref{fig:payload_4_weights}. These payloads are placed at the same place, in the front of the robot, which will generate moment along the pitch axis. Each test for a PLC is repeated five times. The experiment results are depicted as shown in Fig. \ref{fig:converge_plc}. As shown in the figure, when PLC is low, less than 0.4 in this experiment, the identification is able to converge fast. In fact, due to the robot's inertia, the payload has limited influence on the robot. Even with the open-loop force control method, the robot can achieve satisfactory performance. However, given a large PLC, which is yet fully tested on electrically-actuated quadruped robots, the influence of the payload on the robot can not be ignored. It will take a few seconds before the identification converges. As shown in Fig. \ref{fig:converge_plc}, for different PLC, the convergence time is consistently around 2 $s$. The convergence rate can be faster by manually tuning $c, \lambda$, and $M$, in this study the performance is constrained within a safe range, i.e. 2 $s$, to prevent possible overshooting for each robot joint and the potential overturning for the robot.

\begin{figure}
    \centering
    \setlength{\abovecaptionskip}{-0.1 cm}
    \includegraphics[width=3.4 in]{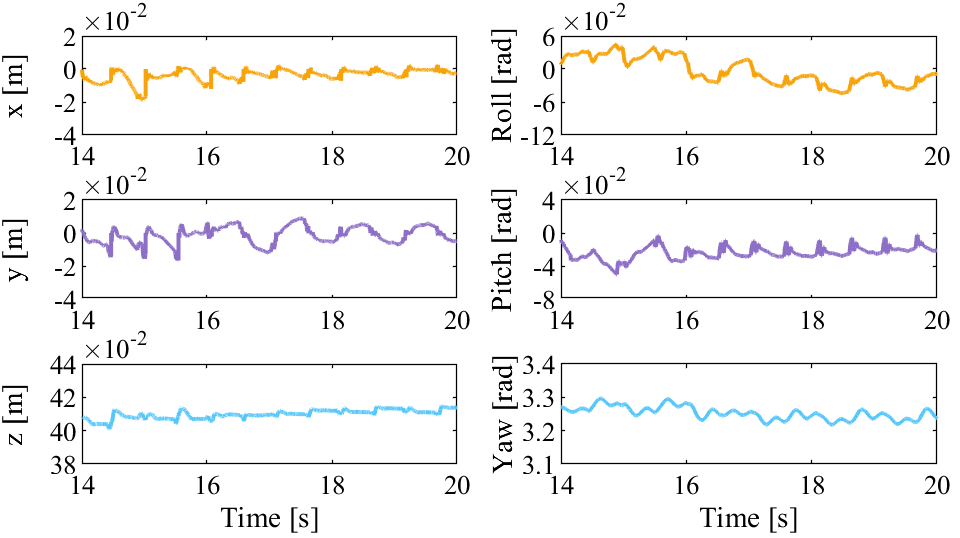}
    \caption{The robot position and orientation in the world coordinate during trotting with the payload.}
    \label{fig:com}
\end{figure}

\begin{figure}
    \centering
    \setlength{\abovecaptionskip}{-0.1 cm}
    \includegraphics[width=3.4 in]{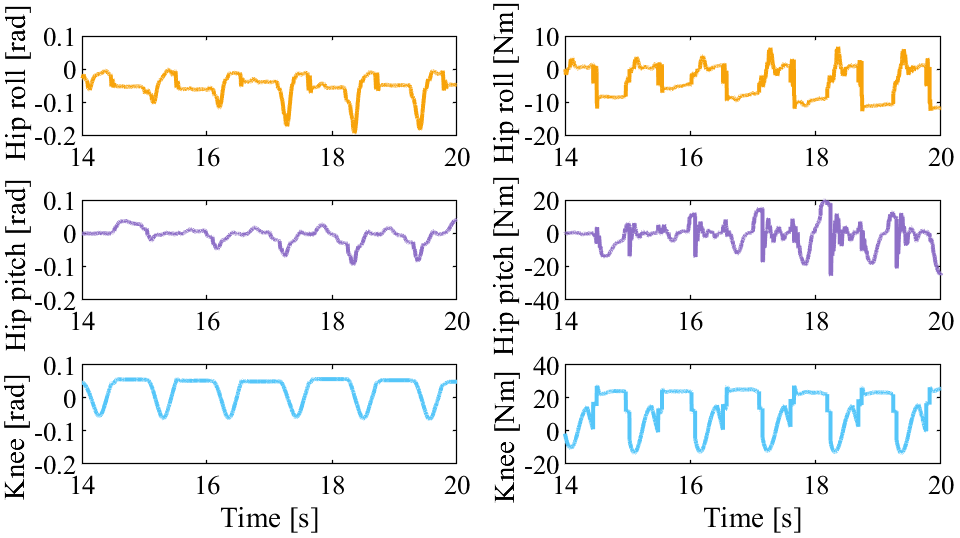}
    \caption{The information of the right front leg of the robot. The left figure is the actual joint position and the right figure is the actual joint torque.}
    \label{fig:joint_torque}
\end{figure}

\subsection{Payload-carrying Trotting Test with ACQL}
In this subsection, the robot is designed to trot with a payload of 50 $kg$, as heavy as the robot's own weight. The experiment screenshots is shown in Fig. \ref{fig:exp_screenshot_plc}. The predefined time of the swing phase and the stance phase are both 0.5 $s$. The results of the experiment are shown in Fig. \ref{fig:com} and Fig .\ref{fig:joint_torque}. The robot is designed to trot in place at a desired robot CoM position (0, 0, 0.41) with the desired orientation (3.25, -0.01, 0), representing the yaw, pitch, and roll. As shown in Fig. \ref{fig:com}, the robot trotting with the position deviation around 0.01 $m$ in each direction and the orientation less than 0.03 $rad$ in each axis. It can be derived that the robot is able to trot with the desired position and orientation even with the payload as heavy as itself. The locomotion performance is not discounted by the payload. Usually, the torque of the knee joint will be two times that of the hip pitch joint if all the robot joints are designed as the articulated ones \cite{kim2019highly}. Benefiting from the prismatic knee joint in our quadruped robot Kirin, the actual torque of the knee joint is similar to that of the hip pitch joint as shown in Fig. \ref{fig:joint_torque}. This means that, with the prismatic knee joint instead of the traditional rotation joint, the robot's ability of payload-carrying is greatly improved.

\section{Conclusion}
This letter presents an online high-payload identification and adaptive control for an electrically-actuated quadruped robot. By the aid of the identified mass and moment of the unknown payload, the quadruped locomotion is able to adapt to the dynamically changing high-payload. In this study, an electrically-actuated quadruped robot for heavy payload-carrying, Kirin, is used the support the tests to verify the effectiveness of the proposed method, ACQL. ACQL is validated for a wide range of PLC from 0.2 to 1.5. The effectiveness of ACQL is also verified in the trotting gait with a payload as heavy as the robot itself. Statistic results show that ACQL is able to converge fast and run efficiently online, which demonstrates that the proposed method is valid for the payload with no matter unknown weight or unknown location. However, at the current stage, ACQL is still limited for the payload with static weight. Therefore, the future work will include dynamic quadruped locomotion control with dynamically changing payload.

\section{Acknowledgement}
We would like to thank Mr. Juntong Su and Mr. Shusheng Ye for their assistance in the experiments.

\bibliographystyle{IEEEtran}
\bibliography{8-mybib}

\end{document}